\documentclass[letterpaper]{article} 
\usepackage{aaai2026}  
\usepackage{times}  
\usepackage{helvet}  
\usepackage{courier}  
\usepackage[hyphens]{url}  
\usepackage{graphicx} 
\usepackage{natbib}  
\usepackage{caption} 
\usepackage{algorithm}
\usepackage{algorithmic}
\usepackage{bm}
\usepackage{subcaption}
\usepackage{booktabs}
\usepackage{makecell}
\usepackage{amsmath}
\usepackage{mathrsfs}
\usepackage{amssymb}
\DeclareMathAlphabet{\mathpzc}{OT1}{pzc}{m}{it}

\usepackage{newfloat}
\usepackage{listings}
\DeclareCaptionStyle{ruled}{labelfont=normalfont,labelsep=colon,strut=off} 
\floatstyle{ruled}
\newfloat{listing}{tb}{lst}{}
\floatname{listing}{Listing}
\nocopyright

\title{LLM-Driven Multi-Turn Task-Oriented Dialogue Synthesis for Realistic Reasoning}
\author{
    Yu Zhu,
    Kai Yang
}
\affiliations{
    Department of Computer Science and Technology, Tongji University
}

\usepackage{bibentry}

\begin{document}

\maketitle

\begin{abstract}
The reasoning capability of large language models (LLMs), defined as their ability to analyze, infer, and make decisions based on input information, is essential for building intelligent task-oriented dialogue systems. However, existing benchmarks do not sufficiently reflect the complexity of real-world scenarios, which limits their effectiveness in evaluating and enhancing LLM reasoning in practical contexts. Many current reasoning datasets are overly simplistic and abstract, often disconnected from realistic task flows, domain constraints, and operational rules, making it difficult to effectively evaluate LLMs' logical reasoning ability. In addition, data contamination from pretraining corpora undermines the reliability of evaluation results, and traditional crowdsourcing methods for dataset construction are labor-intensive and difficult to scale.
To address these challenges, we propose a LLM-driven framework for synthesizing multi-turn, task-oriented dialogues grounded in realistic reasoning scenarios, leveraging trilevel optimization to enhance dialogue quality. Our method generates dialogues grounded in authentic task scenarios, enriched with real-world information, and exhibiting strong contextual coherence. Corresponding reasoning tasks are carefully designed around these dialogues and iteratively refined to continuously improve the tasks’ quality and challenge. The resulting dataset serves as a valuable benchmark for assessing and advancing the realistic logical reasoning capabilities of LLMs. Experimental results show that our synthetic data–based reasoning tasks introduce non-trivial reasoning challenges and provide meaningful support for improving the reasoning capabilities of LLMs.
\end{abstract}

\begin{links}
    \link{Datasets}{https://github.com/adventureoflingke/RealReasoning/}
\end{links}

\section{Introduction}
With the rapid development of large language models, natural language processing has entered a new era\citep{achiam2023gpt4,ouyang2022training,wei2022emergent}. Large-scale pre-trained language models, empowered by transformer architectures and massive parameter sizes, have achieved remarkable performance across a wide range of NLP tasks, including text classification, machine translation, question answering, and summarization\citep{gao2025llm,kernycky2024evaluating,li2024pretrained}. These models often achieve near-human or even super-human performance on benchmark datasets, significantly pushing the boundaries of what was previously thought possible\citep{chang2024survey}. However, despite these impressive achievements, when faced with logical reasoning tasks in realistic scenarios, their performance remains unsatisfactory\citep{li2024llms}. Therefore, to enhance the realistic logical reasoning capabilities of LLMs, acquiring high-quality data grounded in realistic environments has emerged as a critical direction for future research.

Reasoning tasks serve as a key means to evaluate a model’s reasoning capabilities, and the quality of the datasets used in these tasks has a decisive impact on research outcomes\citep{lu2023survey,yu2024natural}. As models grow increasingly sophisticated\citep{huang2023towards}, the need for datasets that accurately measure reasoning capability becomes more pressing. Although multiple datasets exist for evaluating reasoning ability\citep{chen2022convfinqa,zhang2021noahqa}, they suffer from notable limitations. First, the tasks used in these datasets are often relatively simple and abstract, present reasoning tasks in a highly intuitive manner, rarely accounting for factors appear in realistic scenarios. Logical reasoning problems that involve realistic scenarios, integrate multi-source information, and are closely tied to business rules are seldom covered, such problems typically require models to maintain coherence across long sequences, and apply context-sensitive constraints. Apart from it, realistic scenarios often involve privacy concerns for both individuals and organizations\citep{yao2024survey}. The commercial value associated with real data further complicates its collection and public release, leading to an increasing scarcity of such data in public. Moreover, existing evaluation benchmarks suffer from data contamination due to the overlap between pretraining data and test sets, undermining the reliability of performance assessments\citep{deng2024unveiling,magar2022data}. Traditional approaches mainly rely on crowd-sourcing for dataset construction, and thus are inherently limited in addressing such issues. Existing works have also explored agent-based approaches for constructing dialogue datasets, which rely on pre-specified structured task information to generate dialogues\citep{rastogi2020towards,shah2018bootstrapping}. While these datasets provide important benchmarks for evaluating model capabilities in information extraction and task completion, their generation processes are highly dependent on predefined schemas and flows, thus limiting their ability to reflect the complex reasoning exhibited in real-world scenarios. Moreover, in the evaluation of synthetic dialogue quality, the process of exploring and iteratively refining evaluation metrics is often time-consuming and resource-intensive, and there remains a lack of efficient and high-quality evaluation methodologies.
 Therefore, developing methods to construct logical reasoning datasets grounded in real-world scenarios in a more efficient manner plays a crucial role in advancing the development and application of reasoning models in practical settings.

To address these challenges, we propose a novel approach that leverages large language model agents to simulate user behaviors based on realistic scenarios. Our goal is to overcome the limitations of traditional dataset construction methods, such as lack of realism and data contamination, by generating high-fidelity interactions in a controlled and scalable manner. Our method generates action sequences that closely resemble realistic interactions. These sequences are constructed by modeling the decision-making patterns of users during realistic scenarios, which ensures that the user agent possesses unique and realistic action memory aligned with realistic scenarios. Based on these sequences, user agents and assistant agents engage in dialogue, producing multi-turn task-oriented dialogues grounded in authentic real-world scenarios by prompting the model. Furthermore, we propose a trilevel optimization framework that jointly optimizes the evaluation metric, multi-turn dialogue prompt, and single-turn dialogue prompt, enabling automatic evaluation and iterative refinement of synthesized dialogue quality. This framework establishes an end-to-end pipeline from dialogue generation to evaluation, effectively supporting the production of high-quality synthetic dialogues.
The contributions of this paper can be summarized as follows.

\begin{itemize}
\item We formalize dialogue generation as a trilevel optimization problem that jointly optimizes prompt design and evaluation criteria, enabling the automatic generation and iterative refinement of high-quality synthetic dialogues. By leveraging the generative and role-playing capabilities of LLM agents, our approach constructs task-oriented, logical reasoning-intensive dialogue data grounded in realistic information and aligned with scenario-specific rules.

\item We present a multi-turn task-oriented dialogue dataset specifically designed to evaluate models’ logical reasoning capabilities in realistic scenarios. Each instance in the dataset consists of a reasoning question derived from a realistic multi-turn dialogue context, accompanied by a corresponding label that has been manually annotated and verified. While such task-oriented datasets remain scarce, they play a pivotal role in advancing the development and deployment of reasoning models in practical settings, thereby offering significant value for research.

\item Our experiments are evaluated on multiple baseline models and the results show that there is still significant room for improvement in current models when handling such reasoning tasks. This indicates that our dataset not only provides a challenging evaluation platform for existing models but also serves as a valuable resource for future research. Therefore, the proposed method has the potential to drive advancements in the reasoning capabilities of LLMs.
\end{itemize}

\section{Related work}
\subsection{LLM-based agents}
Due to their impressive planning and reasoning capabilities, large language models (LLMs) are increasingly being used as the foundation for constructing intelligent agents\citep{kojima2022large,li2023systematic,nguyen2023cof,xi2025rise}. Specifically, LLMs serve as the central controller of these agents, enabling them to perceive external information and take actions accordingly\citep{shinn2023reflexion,yao2023react}. LLM-based agents possess strong interaction capabilities with their environments. Moreover, thanks to their pre-training on massive corpora, LLMs have accumulated extensive knowledge across diverse domains, allowing them to generalize and transfer across tasks without task-specific fine-tuning\citep{chung2023instrodtods,jiao2025pr,wang2020dual}. 

With recent advances in related research, multi-agent systems built upon LLMs have achieved notable success in solving complex problems and simulating realistic scenarios\citep{guo2024large}. LLM-based agents can communicate with each other, enabling both cooperation and competition\citep{he2023lego,hongmetagpt,islam2024mapcoder,liu2025nested,qian2024chatdev,zhang2024proagent}. By specializing agents in terms of roles and skills, LLM-based multi-agent systems can be assigned distinct capabilities and responsibilities, thereby enhancing performance on specific tasks through division of labor. The collaboration among agents with complementary roles leads to superior overall system performance, demonstrating the potential of LLM-based multi-agent architectures in modeling complex, realistic behaviors\citep{owoicho2023exploiting,park2023generative}. Based on existing research findings, this paper leverages the division of labor and collaboration among multiple agents to enhance the simulation of real-world scenarios. By assigning a specific agent to focus on a single task, the quality of the generated data is further improved.

\subsection{Multi-trun dialogue data generation}
Training machines to understand natural language and interact with humans is a challenging yet fundamental task in artificial intelligence. With the rapid advancement of deep learning techniques, especially the widespread adoption of pre-trained language models (PLMs), various dialogue systems have been proposed that are increasingly capable of engaging in multi-turn conversations\citep{algherairy2024review,ham2020end,kulhanek2021augpt,lin2020mintl,xu2021topic}.
Compared to traditional text-based reading comprehension tasks, multi-turn dialogues are more closely aligned with spoken language and exhibit strong interactive characteristics. These dialogues typically involve multiple speakers, diverse intents, and frequent topic shifts, resulting in utterances that are highly dynamic and context-dependent\citep{huang2020challenges,ma2023enhanced,su2019improving}. Performing reasoning tasks in such settings introduces significantly greater complexity than understanding static, single-paragraph texts.

Traditional dialogue datasets are primarily collected and constructed through crowd-sourcing\citep{budzianowski2018multiwoz,feng2023mmdialog,quan2020risawoz,serban2018survey,zhu2020crosswoz} . However, this approach is costly and has limited scalability. To address these challenges, synthetic data generation methods have gained development to some extent\citep{anabytavor2020do,ding2020daga,meng2022generating,ye2022zerogen}. In particular, for dialogue-formatted data, common approaches include enhancing existing datasets\citep{chen2021simple,li2023s2m} or directly generating data based on specific tasks, often leveraging external knowledge to produce multi-turn conversational structures\citep{abdullin2023synthetic,kim2022generating,qiu2024smile}. Building upon existing research, this paper focuses on generating multi-turn task-oriented dialogue, aiming to embed necessary reasoning information within the dialogue structure, thereby enriching the complexity and practical value of the generated dialogues.

\subsection{Trilevel optimization}
Trilevel optimization has emerged as a powerful framework for modeling complex learning problems in which objectives and constraints are interdependent across multiple layers of control\citep{jiao2025federated,jiaoasynchronous,yang2008distributed}, and it has already found numerous applications in machine learning. When gradient information is available, researchers typically adopt first-order gradient-based methods; this setting has been extensively studied, with applications including neural architecture search\citep{guo2020meets,jiao2022timeautoad}, hyperparameter optimization\citep{raghu2021meta,sato2021gradient}, data reweighting\citep{garg2022learning}, OOD generalization\citep{jian2024tri} and distributed optimization\citep{jiao2024provably,liu2025argus}. In scenarios where gradients are unavailable, zeroth-order information is used to solve
 trilevel learning problems, and a limited body of work has recently emerged in this direction\citep{jiaodtzo}. In this paper, we make the first attempt to introduce a zeroth-order trilevel optimization framework to dialogue generation, and we specifically tailor the optimization to the structure of the loss function to more accurately evaluate dialogue quality .

\section{Method}
\subsection{Problem Definition}
This paper presents a method aimed at generating multi-turn, task-oriented dialogues to evaluate reasoning capabilities in realistic scenarios. In this setup, a user agent interacts with a assistant agent through dialogue to accomplish specific tasks in practical domains. Specifically, we denote $x$ as a particular realistic scenario, $\mathcal{F},\mathcal{G}$ as the user agent and the assistant agent, respectively. The resulting multi-turn dialogue is represented as $\bm{Y}=\{\bm{y}_{1}^{\mathcal{F}},\bm{y}_{1}^{\mathcal{G}},...,\bm{y}_{i}^{\mathcal{F}},\bm{y}_{i}^{\mathcal{G}},...,\bm{y}_{n}^{\mathcal{F}},\bm{y}_{n}^{\mathcal{G}}\}$, where $n$ denotes the number of dialogue rounds and $\bm{y}_{i}^{\mathcal{F}},\bm{y}_{i}^{\mathcal{G}}$ denote the utterances generated by the user agent and the assistant agent in the 
$i$-th turn, respectively.
\subsection{Framework overview}
\begin{figure}
  \centering
  \includegraphics[width=0.45\textwidth]{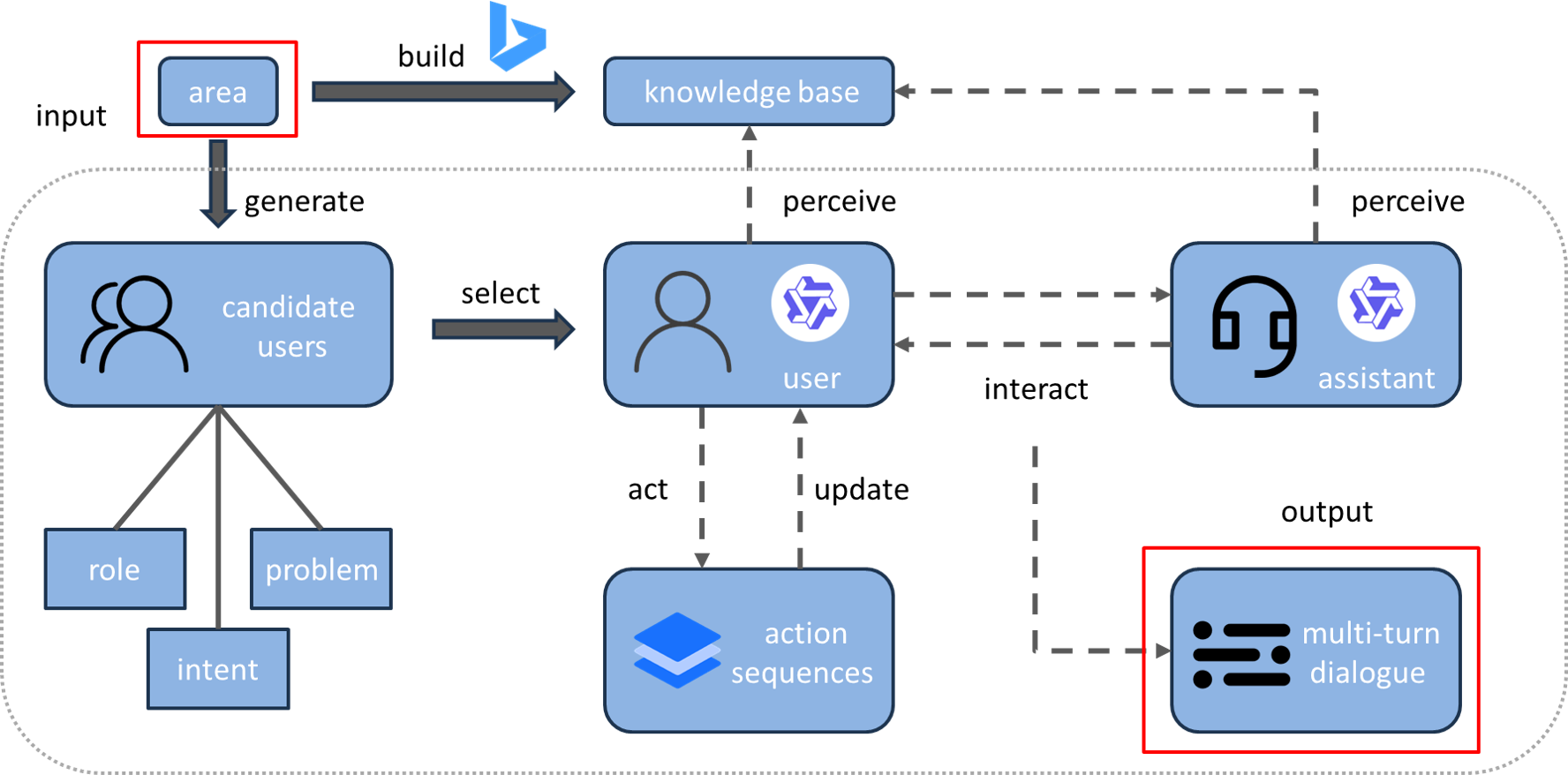}
  \caption{Schematic of the synthetic generation process.}
  \label{fig_overll}
\end{figure}

To evaluate the reasoning capabilities of LLMs, this study aims to simulate multi-turn task-oriented dialogues in realistic scenarios containing implicit information that can only be uncovered through logical reasoning. We propose a framework for generating multi-turn task-oriented dialogues driven by LLM-based agents. Given a realistic scenario, our framework inputs its description into an LLM-based user generator to get user agent, user agent observe the environment and perform actions. After each action, the user agent records its experience, and this process repeats to generate a sequence of actions.
Once a complete action sequence is formed, the user agent engages in a turn-by-turn dialogue with an assistant agent. The user agent asks questions based on the action sequence and scenario context, and the assistant agent reasons about these questions using externally perceived information and provides responses. The generation process is shown in the figure \ref{fig_overll}.

After dialogue generation, we automatically evaluate and improve dialogue quality within a trilevel optimization framework by automatically searching for more accurate evaluation metrics and refining the dialogue prompts of the agents, thereby continuously enhancing the quality of the generated dialogues.

\subsection{User generator}
Generally, different users will have different roles, goals and problems to solve when faced with a particular realistic scenario. In the first step, we employ a LLM, referred to as the user generator, dedicated to generating users' intents and potential issues by prompting it. After generating the user's intent and problem, the created user profile is placed into a pool of candidate users, from which they can be selected for subsequent processes. By constructing various user roles, intents and problems in advance, user agents can focus on taking actions without having to generate their own intents and related content, thereby enhancing the performance of the user agents. The generation process of the condidate user by the user generator is formulated as Eq. (\ref{eq1}):
\begin{equation}
    \bm{U}=\mathcal{H}(x,p_{1}),
    \label{eq1}
\end{equation}
where $p_{1}$ is the prompt that instructs the LLM to act as the user generator, $\mathcal{H}$ represents the LLM used to generate candidate users, $\bm{U}=\{\bm{u}_{1},\bm{u}_{2},...,\bm{u}_{i},...,\bm{u}_{m}\}$ represents the generated candidate users, $\bm{u}_{i}$ represents the $i$-th candidate user and $m$ represents the number of candidate users.
\subsection{Simulated actions of the user agent}
After the generation of candidate users, a user is selected from the candidate pool and the user agent will act as the selected user, adopting its role, intent, problems. The user agent takes actions based on the observed external information and its existing memory. Initially, the user’s memory consists of its intent and associated problems. External information is collected using tools such as search engines, with keywords derived from the input realistic scenario, and is organized into documents to influence the user’s action.

Following each action round, the user reviews its action and integrates it with its existing memory to form an updated memory state. By iteratively performing this process, a complete sequence of actions is generated, which simulates the action trajectory a real user might follow in a specific realistic scenario. The process of generating action sequence is formulated as Eq. (\ref{eq2}):
\begin{equation}
    \bm{a_{i}}=\mathcal{F}(x,\bm{A}_{i-1},\bm{K},p_{2}),
    \label{eq2}
\end{equation}
where $\bm{A}_{i-1}=\{a_{1},a_{2},...,a_{i-1}\}$ represents the action sequence consisting of the first $i-1$ actions, $\bm{K}$ represents the external information collected by tools based on the input realistic scenario, $p_{2}$ represents the prompt used to control the user agent to take actions and $\bm{a_{i}}$ represents the action taken by the user agent in $i$-th action round. 

By constructing a sequence of user actions, the user agent acquires stable and well-defined memory content. This enables the agent to exhibit consistent dialogue topics and maintain coherent intent throughout its interaction with the assistant agent, thereby enhancing the task-oriented nature of the dialogue.
Through the incorporation of conditional information stored in memory, the resulting dialogues naturally embed implicit logical structures that require logical reasoning to uncover. Based on these inferred conclusions, we further derive corresponding questions and labels, which can be used for downstream evaluation tasks to evaluate the models' capability to perform realistic reasoning.


\subsection{Interaction between the user agent and the assistant agent}
Providing a fixed prompt into a large language model often results in outputs with limited diversity. The user agent engages in a dialogue with the assistant agent using memories drawn from realistic scenarios after generating its action sequence. This approach ensures that each user agent has distinct experiences, thereby generating more diverse responses during interactions with the assistant agent. Meanwhile, real-world data collected by external tools is incorporated as contextual information, further improving both the diversity and authenticity of the dialogue content. By defining the user agent’s role, intent, and problem context, generating an action sequence, and providing prompt guidance during the interaction,  task-oriented information is effectively integrated into the dialogue. This facilitates the generation of multi-turn task-oriented dialogues that support realistic reasoning tasks. The interaction process between the user agent and the assistant agent can be represented by Eq. (\ref{eq3}) and Eq. (\ref{eq4}): 
\begin{equation}
    \bm{y}_{i}^\mathcal{F}=\mathcal{F}(x,\bm{A},\bm{Y}_{i-1},\bm{K},p_{3}),
    \label{eq3}
\end{equation}
\begin{equation}
    \bm{y}_{i}^\mathcal{G}=\mathcal{G}(x,\bm{A},\bm{Y}_{i-1},\bm{y}_{i}^{\mathcal{F}},\bm{K},p_{4}),
    \label{eq4}
\end{equation}
where $\bm{A}$ represents the complete action sequence generated by the user agent, $,\bm{Y}_{i-1}$ represents the dialogue history up to the $i-1$-th turn, $p_{3}$ and $p_{4}$ represent the prompt used to control the user agent and the assistant agent to interact with each other, respectively.

\subsection{Trilevel optimization for enhancing dialogue quality}  
After the initial dialogue generation, their quality must be assessed, as effective evaluation is crucial for improvement. However, traditional metrics often struggle to comprehensively capture the multidimensional nature of dialogue quality and heavily rely on reference texts, making them ill-suited for evaluating synthetic dialogues. Meanwhile, human evaluation is costly and highly subjective.
Observing that multi-turn dialogues consist of a sequence of consecutive single-turn exchanges, we note that single-turn dialogues focus on the quality of individual responses and serve as the foundation for high-quality multi-turn interactions, while multi-turn dialogues further emphasize global coherence and naturalness built upon this foundation. These two levels are interdependent, and their joint optimization can be naturally formulated as a bilevel optimization problem.
Within this framework, by introducing a module capable of automatically searching for the metric to evaluate dialogue quality and guide model training, the problem can be further formalized as a trilevel optimization problem. This enables end-to-end automatic generation, evaluation, and iterative refinement of multi-turn dialogues, effectively addressing the limitations of existing approaches.
 The  trilevel optimization problem can be formulated as Eq. (\ref{eq_tri}):
\begin{equation}
    \begin{aligned}
    & \min_{\bm{\omega}} \mathpzc{h}(\bm{\theta}, \bm{\phi}) \\
    & \text{s.t.} \quad \bm{\theta} = \mathop{\arg\min}_{\bm{\theta}'} \ \mathpzc{f}(\bm{\omega}, \bm{\theta}', \bm{\phi}; x) \\
    & \quad \text{s.t.} \quad  \bm{\phi} = \mathop{\arg\min}_{\bm{\phi}'} \ \mathpzc{g}(\bm{\omega}, \bm{\theta}', \bm{\phi}'; x) \\
    & \text{var.} \quad \bm{\omega}, \bm{\theta}, \bm{\phi},
\end{aligned}
    \label{eq_tri}
\end{equation}
where $\mathpzc{h}$ is a dialogue quality scoring function implemented using an expert large language model, and $\bm{\omega}\in \mathbb{R}^{d_{\bm{\omega}}}$ denotes the parameters that define the metric, $\mathpzc{f}$ and $\mathpzc{g}$ are the metrics for evaluating multi-turn and single-turn dialogue quality, respectively, both influenced by $\bm{\omega}$, and $\bm{\theta}\in\mathbb{R}^{d_{\bm{\theta}}}, \bm{\phi}\in\mathbb{R}^{d_{\bm{\phi}}}$ represent the model parameters controlling the prompts used in multi-turn and single-turn dialogue generation.
In addition, we consider that the dialogue generation process involves closed-source black-box LLMs, where first-order information is not accessible. Thus, we employ two-point zeroth-order estimate to approximate the gradient for optimizing $\bm{\theta}, \bm{\phi}$, which can be expressed by Eq. (\ref{eq_zo_theta}) and Eq. (\ref{eq_zo_phi}):
\begin{equation}
    \hat{\nabla}\mathpzc{f}(\bm{\theta}')=\frac{ \mathpzc{f}(\bm{\omega}, \bm{\theta}'+\mu_1\bm{u}_1, \bm{\phi}; x)-\mathpzc{f} (\bm{\omega}, \bm{\theta}', \bm{\phi}; x)}{\mu_1}\bm{u}_1,
    \label{eq_zo_theta}
\end{equation}

\begin{equation}
    \hat{\nabla}\mathpzc{g}(\bm{\phi}')=\frac{\mathpzc{g}(\bm{\omega},\bm{\theta}', \bm{\phi}'+\mu_2\bm{u}_2; x)-\mathpzc{g} (\bm{\omega}, \bm{\theta}', \bm{\phi}'; x)}{\mu_2}\bm{u}_2,
    \label{eq_zo_phi}
\end{equation}
where $\bm{u}_1,\bm{u}_2\sim \mathcal{N}(0,\mathbf{I})$ are  random direction vectors and $\mu_1,\mu_2$ are perturbation radii.

To optimize the metric-defining parameters $\bm{\omega}$ , we cast the evaluation metric as a differentiable-loss-like function. However, due to its discrete structure, gradient-based optimization is infeasible, instead, we use an evolutionary algorithm.
Specifically, each candidate loss function is represented as a tree-structured computational graph: the root outputs the final score, internal nodes are operators, and leaves are base dialogue quality metrics (e.g., distinct-n), whose combination is governed by $\bm{\omega}$. The graph is initialized by recursively expanding from the root—sampling operators and adding children according to their arity until at least one root-to-leaf path reaches depth $d$, ensuring sufficient complexity.
After initializing the loss function, we optimize the prompt-control parameters $\bm{\theta}, \bm{\phi}$ based on this loss function and obtain the corresponding dialogue quality score. In each evolutionary iteration, we select the loss function that yields the highest average dialogue quality as the parent, and generate offspring using one of the following three strategies:
\begin{itemize}
\item \textbf{Copy}: Copy the parent’s computational graph.
\item \textbf{Re-initialization}: Generate a new computational graph according to the initialization procedure described above.
\item \textbf{Mutation}: Apply a mutation operation to the parent’s computational graph to produce an offspring.
\end{itemize}
The candidate mutation operations include the following:
\begin{itemize}
\item \textbf{Insertion}:Insert a random operator between a non-root node and its parent, adding a leaf if arity $>$ 1.
\item \textbf{Deletion}: Select an intermediate node for removal, then connect one of its children to its former parent.
\item \textbf{Replacement}: Swap a node’s operator, adjusting children to match the new arity.
\end{itemize}
The evolutionary framework enables efficient exploration of the discrete space of loss functions while leveraging downstream dialogue quality feedback to guide the search.

\section{Dataset}
Based on the aforementioned data generation method, we constructed a dataset named RealReasoning, aimed at evaluating the logical reasoning capabilities of LLMs in real-world scenarios. The dataset was built following the framework proposed in Section 3.1. In this study, we employed Qwen-Max as the large language model to drive the agents, used it to generate the identities of user agents, and simulate the interactions between user agents and assistant agents. The RealReasoning dataset contains 500 multi-turn task-oriented dialogues, all generated through the aforementioned framework. For each dialogue, we first manually design a logical reasoning question based on its content and then annotate a ground-truth label for the question.

\begin{figure*}
    \centering
    \begin{subfigure}[t]{0.315\textwidth}
        \includegraphics[width=\textwidth]{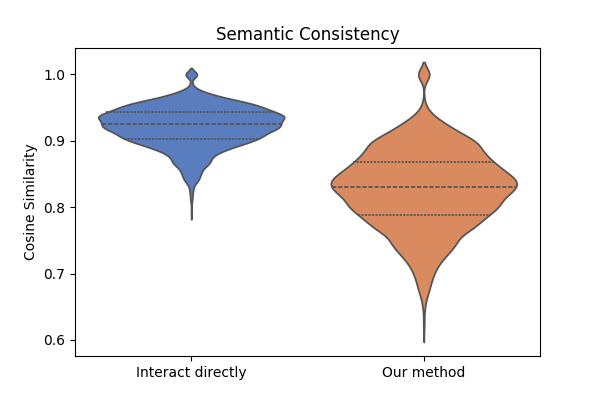}  
        \caption{Cosine similarity of dialogues generated by two distinct approaches.}
        \label{fig2,a}
    \end{subfigure}
    \hfill  
    \begin{subfigure}[t]{0.315\textwidth}
        \includegraphics[width=\textwidth]{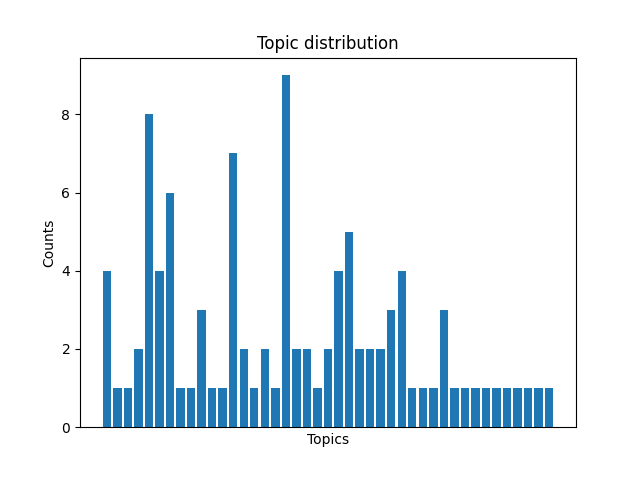}  
        \caption{Topic distribution of dialogues generated by interacting directly.}
        \label{fig2,b}
    \end{subfigure}
    \hfill  
    \begin{subfigure}[t]{0.315\textwidth}
        \includegraphics[width=\textwidth]{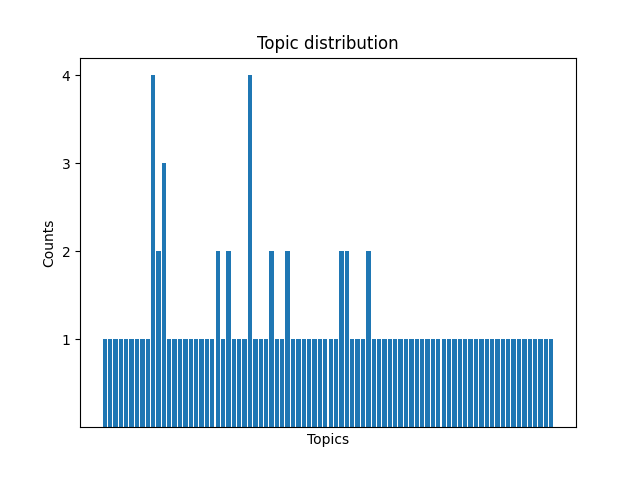}  
        \caption{Topic distribution of dialogues generated by our method.}
        \label{fig2,c}
    \end{subfigure}
    
    \caption{The quality of generated dialogues.}
    \label{fig2}
\end{figure*}

\subsection{Dataset description and statistical analysis}
In the construction of the external knowledge source, we utilize the Bing search engine to retrieve relevant external knowledge. For each search, keywords are extracted from a predefined lexicon that covers a wide range of scenarios requiring logical reasoning, including mathematical and commonsense reasoning. The plain text content from the retrieved web pages is then used to build the external knowledge base related to the scenario.

Additionally, we provide statistical details regarding the dataset, as is shown in Table \ref{tab:realreasoning_stats}. 

\begin{table}[h]
    \centering
    
    \begin{tabular}{c c}
        \toprule
        \textbf{Metric} & \textbf{RealReasoning} \\
        \midrule
        Dialogues & 500 \\
        Total turns & 2398 \\
        Avg. turns per dialogue & 4.796 \\
        Avg. tokens per turn & 69.456 \\
        Unique tokens/Total tokens & 0.024 \\
        \bottomrule
    \end{tabular}
    \caption{RealReasoning Dataset Statistics}
    \label{tab:realreasoning_stats}
\end{table}

\subsection{Reasoning task construction and human annotation}
We manually designed logical reasoning questions based on the dialogues generated by our method and annotated corresponding ground-truth labels, as the dataset inherently poses a certain level of reasoning difficulty. Even state-of-the-art LLMs struggle to provide consistently accurate answers with high reliability. Therefore, it poses certain challenges to achieve fully automated label annotation using LLMs and manual annotation continues to be an essential component in the process. In other words, if current LLMs are able to provide highly accurate and consistently reliable answers, it suggests that the generated reasoning tasks are too simplistic to effectively evaluate the models' reasoning capabilities.

We carefully read and analyzed the multi-turn task-oriented dialogues and designed questions based on the task-oriented information they contain. Logical reasoning tasks refer to tasks that require deriving new conclusions or unknown information from known information or premises through logic and rules. These tasks cannot be solved simply by extracting textual information. Specifically, we constructed two main categories of logical reasoning tasks — math word reasoning and common-sense reasoning — based on the dialogue content. Math word problem reasoning involves scenarios that describe characters, entities, and quantities. These problems typically consist of multi-step arithmetic operations and assess a model’s abilities in language comprehension, semantic parsing, and mathematical reasoning. The label for a math word reasoning task is an exact number that must be obtained through multi-step calculations and cannot be directly derived. In contrast, common-sense reasoning requires synthesizing multiple pieces of information, combining explicitly stated knowledge from the text with implicit common-sense knowledge from the model to yield the final answer. Common-sense reasoning tasks use binary classification labels. For each data, we provided question prompts and corresponding labels for the associated multi-turn dialogues. In the dataset, common-sense reasoning tasks account for 41\%, while math word reasoning tasks account for 59\%.

\subsection{Iterative updating of the dataset}
To further improve the quality of the data in the dataset and ensure that it adequately evaluates the reasoning capabilities of models, this paper adopts an iterative optimization approach, building upon the existing data in the dataset. Given a dataset $\mathcal{D}=\{(\bm{Y}^{i},\bm{P}^i,l^{i})\}_{i=1}^{n}$ constructed using the aforementioned method, where $Y^{i}$ represents the $i$-th multi-turn dialogue, $\bm{P}^i$ represents the problem derived from the $i$-th dialogue, $l^{i}$ is the corresponding label for the problem and $n$ is the current size of the dataset. To implement the iterative optimization process, we introduce a knowledge base of difficult problems, denoted as $\bm{K}_{\bm{P}}$, which is used to store challenging problems identified from the dataset.

For each data instance $(\bm{Y}^{i},\bm{P}^i,l^{i})$ in the RealReasoning dataset, where $i\in [1,n]$, we perform reasoning using $k$ reasoning models $\{\mathcal{M}_i\}_{i=1}^k$ and obtain the corresponding results. The average accuracy of solutions for each data instance is denoted as $acc^i$, forming a data quadruplet $(\bm{Y}^{i},\bm{P}^i,l^{i},acc^i)$. Instances solved correctly by all reasoning models, in other words, data instances with $acc^i=1$ are considered as easy problems. Instances with $acc^i<\phi$ are considered as difficult problems, where $\phi$ is a predefined threshold currently set to 0.5. Each difficult problem is then added to the difficult problem knowledge base $\bm{K}_{\bm{P}}$.

After updating the knowledge base, a problem generation model $\mathcal{I}$ is utilized to generate new problems based on existing dialogues $\bm{Y}^{j}$, with reference to the difficult problems stored in $\bm{K}_{\bm{P}}$. The new problem is generated based on both, which can be formulated as eq(\ref{eq6}):
\begin{equation}
    \hat{\bm{P}}=\mathcal{I}(\bm{Y}^{j},\bm{P}^{k}),
    \label{eq6}
\end{equation}
where $\bm{P}^{k}$ is a difficult problem sampled from $\bm{K}_{\bm{P}}$, and $\hat{\bm{P}}$ represents the newly generated problem. To maximize the proportion of high-quality data in the dataset, we select $\bm{Y}^{j},\bm{P}^{k}$ according to specific conditions. Specifically, for all data instances with $acc^j=1$, the corresponding dialogue data $\bm{Y}^{j}$ is provided. After confirming the multi-turn dialogue, the cosine similarity is used to find the difficult problem in the difficult problem knowledge base that has the highest similarity to this dialogue, which can be formulated as eq(\ref{eq7}),eq(\ref{eq5}):
\begin{equation}
    k = \arg\max_{k \in [1, m]} \text{CosineSimilarity}(\bm{Y}^j, \bm{P}^k),
    \label{eq7}
\end{equation}
\begin{equation}
    \text{CosineSimilarity}(\bm{Y}^j,\bm{P}^k)=\frac{\bm{v}_{\bm{Y}^{j}}\cdot \bm{v}_{\bm{P}^{k}}}{\|v_{\bm{Y}_{j}}\| \|\bm{v}_{\bm{P}_{k}}\|},
    \label{eq5}
\end{equation}
where m is the total number of problems in $\bm{K}_{p}$, $\bm{v}_{\bm{Y}_{j}}$, $\bm{v}_{\bm{P}_{k}}$ represent the text embeddings of the dialogue $\bm{Y}_{j}$ and the difficult problem $\bm{P}_{k}$.

\begin{figure}
  \centering
  \includegraphics[width=0.45\textwidth]{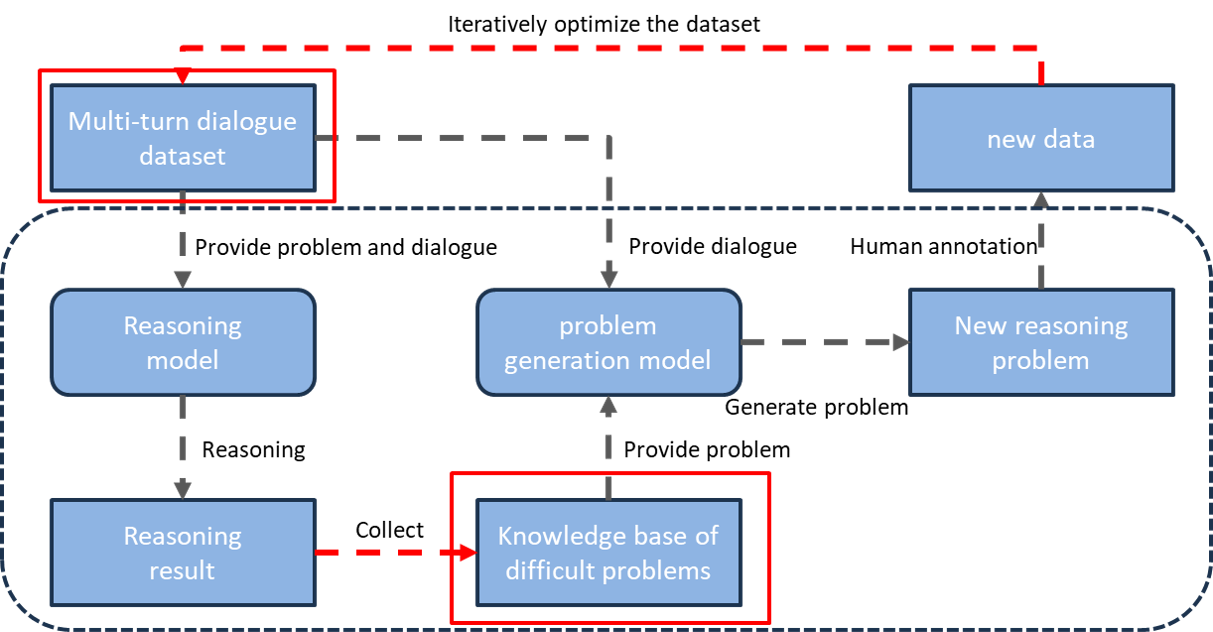}
  \caption{Iterative updating of the dataset.}
  \label{fig_ite}
\end{figure}

Following the problem generation phase, automatic label generation and human verification are conducted to form a new data triplet $(\bm{Y}^{j},\hat{\bm{P}},\hat{l})$, which is then added to the dataset $\mathcal{D}$ to replace the data instance with $acc^j=1$. Through this iterative process, both the dataset $\mathcal{D}$ and the difficult problem knowledge base $\bm{K}_{\bm{P}}$ are iteratively updated. This process will be repeated continuously until there are no more data instance with simple problem. The overall workflow is illustrated in the figure \ref{fig_ite}.
We utilized the Qwen3-235B-Instruct as the problem generation model and employed Qwen-Plus, Qwen-Turbo and DeepSeek-R1-Distill-Qwen-1.5B as the reasoning models. For text embedding, we used thenlper/gte-large-zh. Through this process, we conducted the evaluation and modification of simple problems. After multiple rounds of iteration, the ratio of simple problems dropped from 26.6\% to 0.6\%, effectively enhancing the quality of problems in the dataset, increasing the ratio of challenging problems, and ensuring that the dataset could provide a higher level of challenge as a benchmark.
\subsection{Dialogue evaluation}
To evaluate the quality of the dialogues generated by the proposed method, we employ the semantic consistency metric and record the topic distribution to assess the similarity between generated contents.

We measure dialogue diversity through semantic similarity. Specifically, we use the thenlper/gte-large-zh as the text embedding model to encode utterances into vector representations and compute the cosine similarity $\text{CosineSimilarity}(\bm{Y}^a,\bm{Y}^b)$ between pairs of dialogues to quantify their semantic similarity.

The distribution of cosine similarity is presented in Figure \ref{fig2,a}. We observe that the dialogues generated by directly interacting the user and assistant agents—without pre-assigning a user identity or generating an action sequence—exhibit significantly higher semantic similarity, with an average cosine similarity of 0.921, compared to 0.827 for dialogues produced by our framework.

Moreover, statistical analysis of topic distributions reveals that our method can still ensure sufficient diversity in conversation topics even with limited input. Specifically, we use "business travel reimbursement" as the designated domain for input and facilitate interact directly between a user agent and a assistant agent based on this keyword, either directly or after actions guided by our method. We record the multi-turn dialogues generated and employ Qwen-Max to summarize the topics of these dialogues. As shown in the figure \ref{fig2,b},\ref{fig2,c}, dialogues generated from direct interaction exhibit significant repetition in topics and contain fewer unique topics. In contrast, dialogues generated using our method show virtually no overlap in topics, demonstrating that varied experiences of user agents influence dialogue topics, thereby making the generated content more realistic and better suited for reasoning tasks. We provide more detailed ablation studies in Appendix A.

To further validate the effectiveness of the trilevel optimization framework in improving dialogue quality, we employ qwen-max as a proxy evaluator to score synthesized dialogues along three established dimensions of dialogue quality: Coherence, Fluency, and Diversity. The model is provided with the task description, scoring criteria, and the generated multi-turn dialogue, and assigns scores on a 1–5 scale. We include an unoptimized baseline without any metric-based optimization to demonstrate the effectiveness of different metrics and compare the impact of different automatic evaluation metrics on dialogue quality. As shown in Table \ref{tab:trilevel_performance}, our approach to improving dialogue quality achieves strong performance across all three dialogue quality evaluation dimensions and attains the highest average score, demonstrating that the proposed trilevel optimization framework is effective in improving dialogue quality.

Overall, our framework enhances dialogue diversity and coherence even under minimal input, enabling scalable and realistic dialogue synthesis.

\begin{table}[ht]
    \centering 
    \begin{tabular}{l c c c c}
        \toprule
        Metric & Coherence & Fluency & Diversity & Avg \\
        \midrule
        Unoptimized & 4.33 & 3.08 & 2.96 & 3.46\\
        distinct-N & \textbf{4.91} & 3.50 & 3.07& 3.82 \\
        TF-IDF & 4.35 & 3.28 & 3.10& 3.57 \\
        Embedding & 4.90 & 3.50 & 3.00& 3.80 \\
        ours & 4.90 & \textbf{4.02} & \textbf{3.12}& \textbf{4.01} \\
        \bottomrule
    \end{tabular}
    \caption{The impact of different metrics on dialogue quality.}
    \label{tab:trilevel_performance}
\end{table}

\begin{table*}[ht]
    \centering 
    \begin{tabular}{l c c c c c}
        \toprule
        Model & \makecell[c]{RealReasoning \\ math word \\reasoning} & \makecell[c]{RealReasoning \\ common-sense \\reasoning} & RealReasoning & GSM8K & CODAH \\
        \midrule
        qwen-plus & 40.6\% & 59.5\% & 48.4\% & 69.1\%& 94.8\% \\
        qwen-plus-thinking & 88.1\% & 74.6\% & 82.6\%  & 94.2\%&95.0\% \\
        qwen-turbo & 31.8\% & 52.6\% & 40.4\%  &69.9\% &90.8\% \\
        qwen-turbo-thinking & 88.1\% & 77.5\% & 83.8\%  &\textbf{94.4}\% &86.3\% \\
        deepseek-r1 & \textbf{89.4}\% & \textbf{85.3}\% & \textbf{87.8}\%  & 94.2\% & \textbf{95.2}\% \\
        deepseek-r1-distill-qwen-32b& 78.3\% & 71.7\% & 75.6\%  & 68.8\%&88.6\% \\
        deepseek-r1-distill-qwen-1.5b& 37.6\% & 40.0\% & 38.6\%  & 61.7\%&35.4\% \\
        deepseek-r1-distill-llama-70b& 48.4\% & 69.2\% & 57.0\% & 64.2\%& 93.5\%\\
        \bottomrule
    \end{tabular}
    \caption{Answer accuracy of different models on datasets.}
    \label{tab:reasoning_performance}
\end{table*}

\section{Experimental results}
Experiments are conducted on the RealReasoning dataset to evaluate the ability of large language models to perform realistic logical reasoning. The experimental results and corresponding analyses are provided below to explore the impact of the RealReasoning dataset and the factors influencing model reasoning performance.
\subsection{Main results}
We conduct reasoning tasks on the synthetic dataset using various LLMs and collect the corresponding outputs. Each model is presented with reasoning tasks together with their associated dialogue contexts. We instruct the model to output only the final result, without including any intermediate reasoning steps or internal thought processes. The generated responses are then compared to the ground truth annotations to evaluate performance. We also employ two publicly available datasets, GSM8K\citep{cobbe2021training} and CODAH\citep{chen2019codah}
, for comparison. As is shown in Table \ref{tab:reasoning_performance}, the result shows that even advanced LLMs, such as qwen-plus, struggle to achieve strong performance on the realistic logical reasoning tasks presented in our dataset.
Meanwhile, models that come with an inherent reasoning process, such as DeepSeek-R1, have demonstrated better performance on tasks that require complex reasoning to solve. 

We primarily used models from the qwen series\citep{yang2024qwen2} and the deepseek series\citep{guo2025deepseek} for our experiments. Among them, qwen-plus, and qwen-turbo models responded to problems under a zero-shot prompting scenario. The remaining models  underwent thorough reasoning and deliberation before outputting their answers, specifically, qwen-plus-thinking and qwen-turbo-thinking refer to invoking qwen-plus and qwen-turbo with `enable\_thinking` set to `True`.
\subsection{Analysis and findings}
\paragraph{Reasoning vs No-reasoning.} Based on the RealReasoning dataset, we conducted zero-shot prompting experiments using the qwen series models. The results indicate that when models are not prompted to engage in reasoning and are instead asked to directly solve reasoning tasks, their performance is suboptimal. Even large-scale models such as qwen-plus achieve only around 48.4\% overall accuracy. Smaller models like qwen-turbo exhibit a more pronounced performance degradation on common-sense reasoning tasks, making them ineffective for realistic logical problem-solving.

Upon further examination of the output, it becomes evident that weaker models tend to mechanically select the first available option when faced with closed-ended reasoning questions. This behavior reflects an inability to deeply understand or effectively analyze specific problem contexts, significantly limiting their performance on realistic logical reasoning tasks.

Conversely, models equipped with mechanisms for reflective reasoning demonstrate significant advantages on tasks requiring multi-step inference or deep comprehension. For instance, the DeepSeek model series leverages extensive reflection to thoroughly analyze problems, leading to more accurate responses. Specifically, DeepSeek-R1 utilizes long chains of reasoning to meticulously explore each question. By virtue of its vast parameter size, DeepSeek-R1 accumulates substantial internal knowledge during pre-training, enabling superior performance on realistic logical reasoning tasks. The performance gap between qwen-turbo-thinking and qwen-turbo clearly demonstrates the effectiveness of a thorough reasoning process in solving reasoning tasks.

\paragraph{Math word reasoning vs Common-sense reasoning.} When addressing two distinct types of reasoning tasks—math word reasoning and common-sense reasoning—the latter exhibits significantly greater difficulty. Even the best-performing model in current experiments, deepseek-r1, achieves only an 85.3\% accuracy rate on common-sense reasoning tasks. This indicates that, for tasks requiring auxiliary internal knowledge embedded within model parameters, it is challenging to substantially improve reasoning accuracy solely through explicit reasoning processes. Current models with strong reasoning capabilities tend to focus more on external information provided as input, lacking sufficient engagement with the implicit internal knowledge contained within their parameters.

In contrast, math word reasoning tasks provide all necessary information externally, enabling even small-parameter models like deepseek-r1-distill-qwen-1.5b to effectively extract and utilize this information to derive correct answers. The characteristics of the math word reasoning tasks lead to a larger performance gap between models with reasoning and those without. 

\paragraph{Summary.} Experimental results demonstrate that, compared to tasks from existing public reasoning datasets that are relatively easier to solve, models still face certain difficulties when solving logical reasoning tasks in real-world scenarios. This highlights the value of the data generation method proposed in this paper for evaluating model reasoning capabilities. 

\section{Conclusion and Future Work}
This study presents a novel framework for generating multi-turn task-oriented dialogues using LLM agents, with a focus on evaluating models' logical reasoning capabilities in realistic scenarios. By leveraging the generative and role-playing strengths of LLM agents, our approach effectively constructs dialogue data that is grounded in real-world contexts. Real-world scenarios help maintain contextual coherence across dialogue turns and inherently increase the difficulty of logical reasoning. In addition, we introduce a newly curated dataset specifically designed to assess model performance on such reasoning tasks. Experimental results demonstrate that current models still face significant challenges in handling these complex reasoning tasks, highlighting the need for further research in this direction.

\paragraph{Limitations.} The annotation of the data generated in this study relies on manual labeling, and expanding the dataset still requires considerable human effort. Moreover, the quality of the generated data can be influenced by the performance of the LLMs used.

We plan to explore two main directions for future work. First, we aim to extend the dialogue generation framework from user-assistant interactions to multi-agent settings by incorporating more sophisticated prompting strategies and memory mechanisms. Second, we intend to investigate automated methods for labeling generated dialogue data, aiming to maintain task difficulty while reducing the reliance on labor-intensive manual annotation processes.

\bibliography{aaai2026}

\clearpage
\newpage

\appendix
\begin{figure*}
    \centering
    \begin{subfigure}[t]{0.5\textwidth}
        \includegraphics[width=\textwidth]{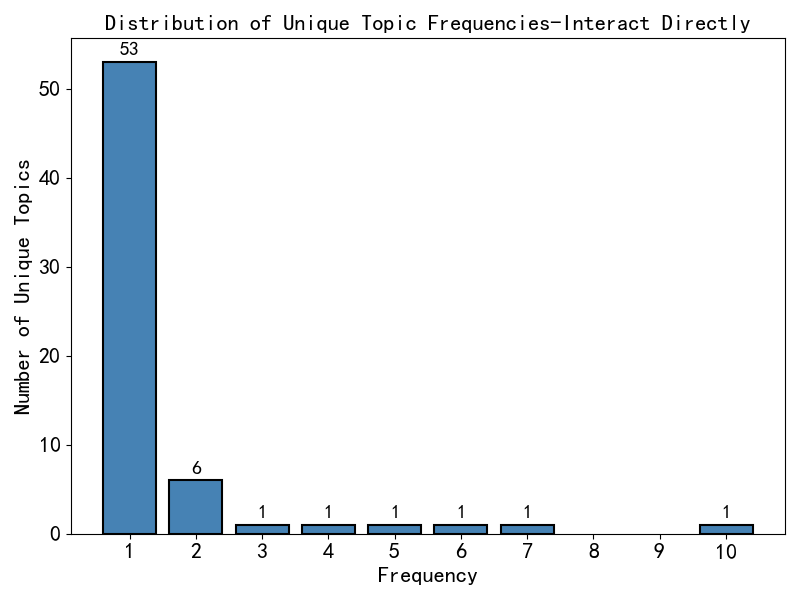}  
        \caption{Distribution of Unique Topic Frequencies-Interact Directly.}
        \label{fig4,a}
    \end{subfigure}
    \begin{subfigure}[t]{0.5\textwidth}
        \includegraphics[width=\textwidth]{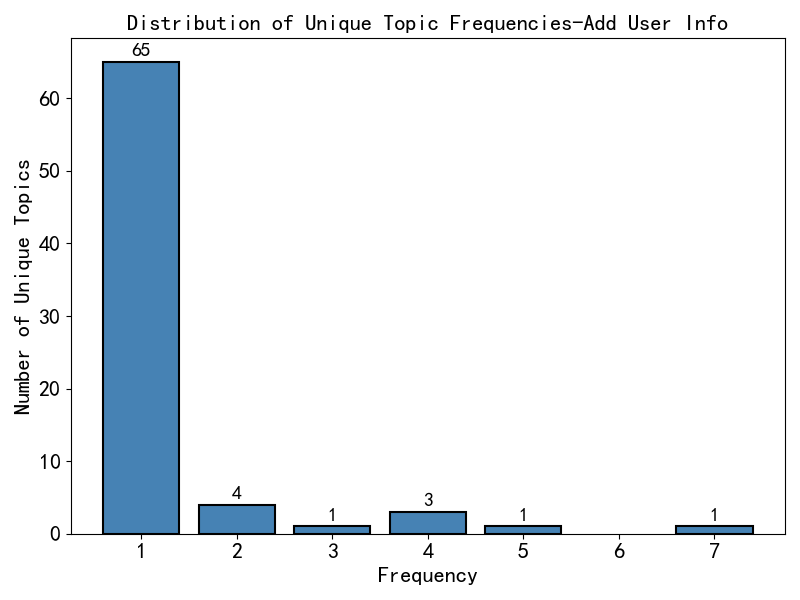}  
        \caption{Distribution of Unique Topic Frequencies-Add User Info.}
        \label{fig4,b}
    \end{subfigure}
    \begin{subfigure}[t]{0.5\textwidth}
        \includegraphics[width=\textwidth]{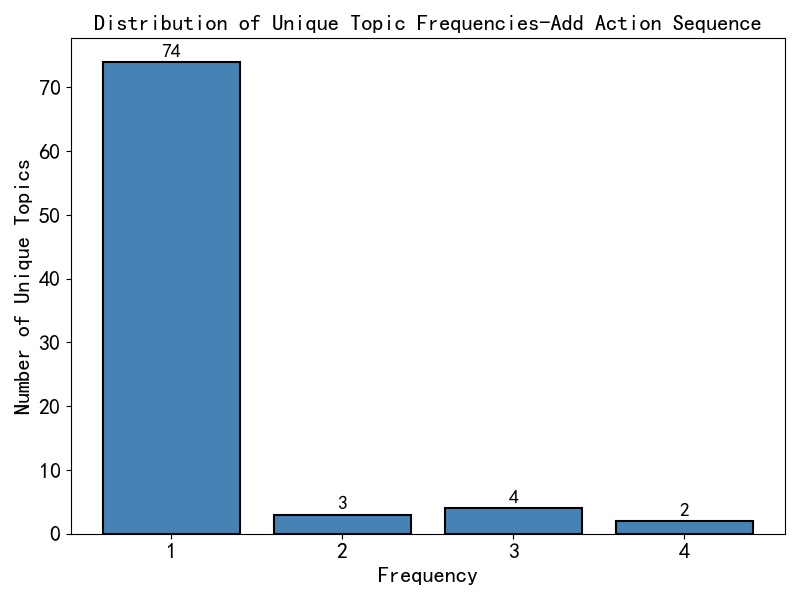}  
        \caption{Distribution of Unique Topic Frequencies-Add Action Sequence.}
        \label{fig4,c}
    \end{subfigure}
    
    \caption{Impact of Ablation Components on Dialogue Topic Distribution.}
    \label{fig4}
\end{figure*}

\begin{figure*}
  \centering
  \includegraphics[width=0.9\textwidth]{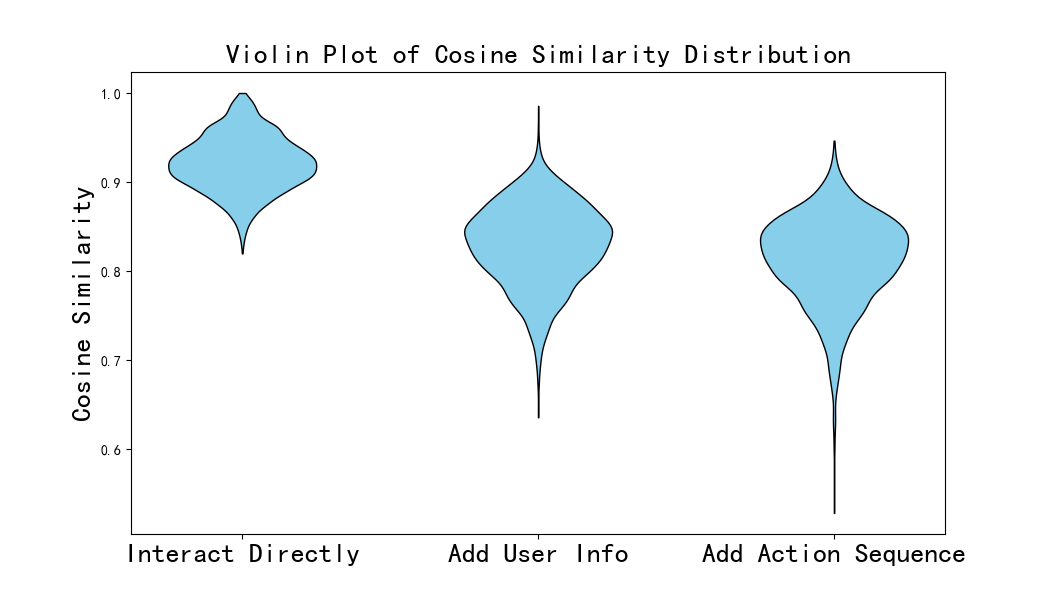}
  \caption{Impact of Ablation Components on Cosine Similarity Distribution.}
  \label{fig_abl_cos}
\end{figure*}

\begin{table*}[htbp]
    \centering
    \caption{Comparison of Multi-turn Dialogue Generation Results with Different Ablation Components Combined}
    \label{tab:dialogue_metrics}
    \begin{tabular}{lccccc}
        \toprule
        \textbf{Method} & \multicolumn{3}{c}{\textbf{Cosine Similarity}} & \multicolumn{2}{c}{\textbf{Topic Diversity and Redundancy}} \\
        \cmidrule(lr){2-4} \cmidrule(lr){5-6}
         & Max & Min & Mean & Unique Topics & Max Repetition \\
        \midrule
        Direct Interaction & 1.00 & 0.820 & 0.921 & 65 & 10 \\
        + User Identity Info & 0.986 & 0.636 & 0.830 & 75 & 7 \\
        + Action Sequence Info & \textbf{0.946}  & \textbf{0.528} & \textbf{0.813} & \textbf{83} & \textbf{4} \\
        \bottomrule
    \end{tabular}
\end{table*}

\section{Ablation Study on Synthetic Dialogues}
This subsection validates the effectiveness of the different modules proposed in this chapter through ablation studies. All dialogues are generated under the same scenario, with user information, action sequence. Each subsequent configuration includes all previously added components, while the baseline, which is direct dialogue generation without any of these components, serves as the reference point. The impact of each component is evaluated by comparing the topic distributions of the generated dialogues and the cosine similarities between their corresponding semantic embeddings. The relevant experimental results are shown in the figure \ref{fig4},\ref{fig_abl_cos} and table \ref{tab:dialogue_metrics}.
As shown in Figure \ref{fig4}, compared to the baseline of direct interaction between user and system agents, equipping the user agent with the modules proposed in this chapter leads to significantly greater diversity in topic distribution. When all modules are incorporated, more than 70\% of generated dialogues exhibit a unique topic, indicating higher inter-dialogue distinctiveness. This suggests that the synthetic dialogues achieve broader coverage of the target scenario and better account for the variety of events likely to occur within it. Moreover, the proportion of unique topics consistently increases as modules are incrementally added, demonstrating that each component positively contributes to enhancing topical diversity.

Figure \ref{fig_abl_cos} further reveals that incorporating the user identity module yields the most pronounced reduction in cosine similarity, while the subsequent addition of the action sequence module also exerts a beneficial effect in further lowering similarity scores.

Finally, Table \ref{tab:dialogue_metrics} shows that under the scene-based direct dialogue setting, the maximum cosine similarity reaches 1.0, indicating the generation of identical dialogue pairs. This reflects a tendency of large language model agents to fall into repetitive response patterns when no auxiliary information is provided, resorting instead to conservative outputs based solely on internal priors. Notably, the introduction of action sequence information substantially reduces the minimum cosine similarity between dialogues.

The statistics on topic diversity and repetition across different methods further validate the effectiveness of our proposed modules. Here, “Unique Topics” denotes the number of distinct dialogue topics covered, and “Max Repetition” refers to the highest frequency of any single topic. Both the user identity and action sequence modules markedly improve topic diversity, thereby contributing positively to overall dialogue quality.

\makeatletter
\@ifundefined{isChecklistMainFile}{
  \newif\ifreproStandalone
  \reproStandalonetrue
}{
  \newif\ifreproStandalone
  \reproStandalonefalse
}
\makeatother

\ifreproStandalone
\documentclass[letterpaper]{article}
\usepackage[submission]{aaai2026}
\setlength{\pdfpagewidth}{8.5in}
\setlength{\pdfpageheight}{11in}
\usepackage{times}
\usepackage{helvet}
\usepackage{courier}
\usepackage{xcolor}
\frenchspacing

\begin{document}
\fi
\setlength{\leftmargini}{20pt}
\makeatletter\def\@listi{\leftmargin\leftmargini \topsep .5em \parsep .5em \itemsep .5em}
\def\@listii{\leftmargin\leftmarginii \labelwidth\leftmarginii \advance\labelwidth-\labelsep \topsep .4em \parsep .4em \itemsep .4em}
\def\@listiii{\leftmargin\leftmarginiii \labelwidth\leftmarginiii \advance\labelwidth-\labelsep \topsep .4em \parsep .4em \itemsep .4em}\makeatother

\setcounter{secnumdepth}{0}
\renewcommand\thesubsection{\arabic{subsection}}
\renewcommand\labelenumi{\thesubsection.\arabic{enumi}}

\newcounter{checksubsection}
\newcounter{checkitem}[checksubsection]

\newcommand{\checksubsection}[1]{%
  \refstepcounter{checksubsection}%
  \paragraph{\arabic{checksubsection}. #1}%
  \setcounter{checkitem}{0}%
}

\newcommand{\checkitem}{%
  \refstepcounter{checkitem}%
  \item[\arabic{checksubsection}.\arabic{checkitem}.]%
}
\newcommand{\question}[2]{\normalcolor\checkitem #1 #2 \color{blue}}
\newcommand{\ifyespoints}[1]{\makebox[0pt][l]{\hspace{-15pt}\normalcolor #1}}

\clearpage
\newpage

\section*{Reproducibility Checklist}

\vspace{1em}
\hrule
\vspace{1em}

\textbf{Instructions for Authors:}

This document outlines key aspects for assessing reproducibility. Please provide your input by editing this \texttt{.tex} file directly.

For each question (that applies), replace the ``Type your response here'' text with your answer.

\vspace{1em}
\noindent
\textbf{Example:} If a question appears as
\begin{center}
\noindent
\begin{minipage}{.9\linewidth}
\ttfamily\raggedright
\string\question \{Proofs of all novel claims are included\} \{(yes/partial/no)\} \\
Type your response here
\end{minipage}
\end{center}
you would change it to:
\begin{center}
\noindent
\begin{minipage}{.9\linewidth}
\ttfamily\raggedright
\string\question \{Proofs of all novel claims are included\} \{(yes/partial/no)\} \\
yes
\end{minipage}
\end{center}
Please make sure to:
\begin{itemize}\setlength{\itemsep}{.1em}
\item Replace ONLY the ``Type your response here'' text and nothing else.
\item Use one of the options listed for that question (e.g., \textbf{yes}, \textbf{no}, \textbf{partial}, or \textbf{NA}).
\item \textbf{Not} modify any other part of the \texttt{\string\question} command or any other lines in this document.\\
\end{itemize}

You can \texttt{\string\input} this .tex file right before \texttt{\string\end\{document\}} of your main file or compile it as a stand-alone document. Check the instructions on your conference's website to see if you will be asked to provide this checklist with your paper or separately.

\vspace{1em}
\hrule
\vspace{1em}


\checksubsection{General Paper Structure}
\begin{itemize}

\question{Includes a conceptual outline and/or pseudocode description of AI methods introduced}{(yes/partial/no/NA)}
yes

\question{Clearly delineates statements that are opinions, hypothesis, and speculation from objective facts and results}{(yes/no)}
yes

\question{Provides well-marked pedagogical references for less-familiar readers to gain background necessary to replicate the paper}{(yes/no)}
yes

\end{itemize}
\checksubsection{Theoretical Contributions}
\begin{itemize}

\question{Does this paper make theoretical contributions?}{(yes/no)}
yes

	\ifyespoints{\vspace{1.2em}If yes, please address the following points:}
        \begin{itemize}
	
	\question{All assumptions and restrictions are stated clearly and formally}{(yes/partial/no)}
	yes

	\question{All novel claims are stated formally (e.g., in theorem statements)}{(yes/partial/no)}
	yes

	\question{Proofs of all novel claims are included}{(yes/partial/no)}
	no

	\question{Proof sketches or intuitions are given for complex and/or novel results}{(yes/partial/no)}
	yes

	\question{Appropriate citations to theoretical tools used are given}{(yes/partial/no)}
	yes

	\question{All theoretical claims are demonstrated empirically to hold}{(yes/partial/no/NA)}
	yes

	\question{All experimental code used to eliminate or disprove claims is included}{(yes/no/NA)}
	no
	
	\end{itemize}
\end{itemize}

\checksubsection{Dataset Usage}
\begin{itemize}

\question{Does this paper rely on one or more datasets?}{(yes/no)}
yes

\ifyespoints{If yes, please address the following points:}
\begin{itemize}

	\question{A motivation is given for why the experiments are conducted on the selected datasets}{(yes/partial/no/NA)}
	yes

	\question{All novel datasets introduced in this paper are included in a data appendix}{(yes/partial/no/NA)}
	yes

	\question{All novel datasets introduced in this paper will be made publicly available upon publication of the paper with a license that allows free usage for research purposes}{(yes/partial/no/NA)}
	yes

	\question{All datasets drawn from the existing literature (potentially including authors' own previously published work) are accompanied by appropriate citations}{(yes/no/NA)}
	NA

	\question{All datasets drawn from the existing literature (potentially including authors' own previously published work) are publicly available}{(yes/partial/no/NA)}
	NA

	\question{All datasets that are not publicly available are described in detail, with explanation why publicly available alternatives are not scientifically satisficing}{(yes/partial/no/NA)}
	NA

\end{itemize}
\end{itemize}

\checksubsection{Computational Experiments}
\begin{itemize}

\question{Does this paper include computational experiments?}{(yes/no)}
yes

\ifyespoints{If yes, please address the following points:}
\begin{itemize}

	\question{This paper states the number and range of values tried per (hyper-) parameter during development of the paper, along with the criterion used for selecting the final parameter setting}{(yes/partial/no/NA)}
	yes

	\question{Any code required for pre-processing data is included in the appendix}{(yes/partial/no)}
	yes

	\question{All source code required for conducting and analyzing the experiments is included in a code appendix}{(yes/partial/no)}
	yes

	\question{All source code required for conducting and analyzing the experiments will be made publicly available upon publication of the paper with a license that allows free usage for research purposes}{(yes/partial/no)}
	yes
        
	\question{All source code implementing new methods have comments detailing the implementation, with references to the paper where each step comes from}{(yes/partial/no)}
	partial

	\question{If an algorithm depends on randomness, then the method used for setting seeds is described in a way sufficient to allow replication of results}{(yes/partial/no/NA)}
	yes

	\question{This paper specifies the computing infrastructure used for running experiments (hardware and software), including GPU/CPU models; amount of memory; operating system; names and versions of relevant software libraries and frameworks}{(yes/partial/no)}
	yes

	\question{This paper formally describes evaluation metrics used and explains the motivation for choosing these metrics}{(yes/partial/no)}
	yes

	\question{This paper states the number of algorithm runs used to compute each reported result}{(yes/no)}
	yes

	\question{Analysis of experiments goes beyond single-dimensional summaries of performance (e.g., average; median) to include measures of variation, confidence, or other distributional information}{(yes/no)}
	yes

	\question{The significance of any improvement or decrease in performance is judged using appropriate statistical tests (e.g., Wilcoxon signed-rank)}{(yes/partial/no)}
	yes

	\question{This paper lists all final (hyper-)parameters used for each model/algorithm in the paper’s experiments}{(yes/partial/no/NA)}
	yes

\end{itemize}
\end{itemize}
\ifreproStandalone
\end{document}
\fi

\end{document}